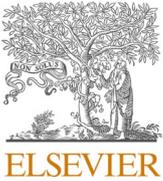
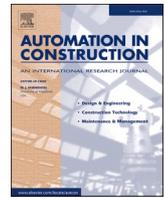

# SL Sensor: An open-source, real-time and robot operating system-based structured light sensor for high accuracy construction robotic applications

Teng Foong Lam [*], Hermann Blum, Roland Siegwart, Abel Gawel

*ETH Zürich D-MAVT, Zurich, Switzerland*

ABSTRACT

High accuracy 3D surface information is required for many construction robotics tasks such as automated cement polishing or robotic plaster spraying. However, consumer-grade depth cameras currently found in the market are not accurate enough for these tasks where millimeter (mm)-level accuracy is required. This paper presents SL Sensor, a structured light sensing solution capable of producing high fidelity point clouds at 5 Hz by leveraging on phase shifting profilometry (PSP) codification techniques. The SL Sensor was compared with to two commercial depth cameras - the Azure Kinect and RealSense L515. Experiments showed that the SL Sensor surpasses the two devices in both precision and accuracy for indoor surface reconstruction applications. Furthermore, to demonstrate SL Sensor's ability to be a structured light sensing research platform for robotic applications, a motion compensation strategy was developed that allows the SL Sensor to operate during linear motion when traditional PSP methods only work when the sensor is static. Field experiments show that the SL Sensor is able to produce highly detailed reconstructions of spray plastered surfaces. The robot operating system (ROS)-based software and a sample hardware build of the SL Sensor are made open-source with the objective to make structured light sensing more accessible to the construction robotics community.

## 1. Introduction

Many automated construction tasks require high accuracy millimeter (mm)-level 3D sensing to ensure successful task execution. One example is automated cement polishing where the robotic system needs to detect uneven bumps on the wall surface to determine the areas that require additional grinding. Another example is robotic plaster spraying [1], where high accuracy sensing is required to perform process monitoring and determine both the location and timing of additional surface treatment [2].

Known sensor technologies for robotic 3D sensing can be roughly sorted into stereoscopic vision, structured light and time-of-flight (ToF) [3]. From these, structured light is the most accurate for surface interaction [4]. Similar to stereoscopic vision, structured light sensors triangulate 3D geometry between camera and projector, but pixels can be matched based on additional information transported in the projection. Therefore, using multiple projections for a single *scan* yields sub-pixel accuracy and reduced measurement uncertainty [5,6], but requires a strategy to handle motion between these projections. Commonly available sensors in robotic applications are therefore limited to ToF or single-projection structured light.

In this paper, we present SL Sensor,[1] an open-source structured light sensor capable of producing detailed 3D scans of sub-mm accuracy in real time. Our sensor integrates with the robot's software and therefore allows to adaptively change the projected pattern as well as the number of projections per scan. While any pattern can be used, we mainly utilise the phase shifting profilometry (PSP) codification technique, which involves shining multiple sinusoidal patterns at the measured surface, to produce a high fidelity reconstruction of the scanned area. PSP techniques usually require the sensor to be static during the entire scanning process and any movement causes artifacts in the resulting scan. A naïve strategy based on our sensor's adaptation capability would already improve over existing sensors: Use high-accuracy PSP projections when the robot is still, and fall back to lower accuracy single-projection techniques [7,8] when the robot is moving. However, to enable high-accuracy sensing for more interactive robotic construction applications, we further propose a motion compensation strategy that allows PSP-based sensing during linear motion. Our motion compensation is

---

based on the specific characteristic of robotic sensing: Because the robot controls its own motion, it can anticipate and compensate for this motion by adapting the projected pattern of the SL Sensor accordingly.

While there are existing open-source implementations of structured light scanning [9,10], it is to the best of our knowledge that SL Sensor is the first open-source structured light scanner project that not only provides the software, but also a documented sample hardware build that can be made using easily available components and open-source electronics. Moreover, SL Sensor is fully compatible with the well-established ROS middleware, enabling easy integration into existing robotic systems.

In summary, the contributions of this paper include

1. The development of an open-source high accuracy structured light sensor that integrates with robotic software to change the projection pattern online, made available in the form of open-source hardware schematics and accompanying software
2. A novel motion compensation strategy for PSP that enables the aforementioned structured light sensor to scan while in linear motion.

## 2. Method

We first give some background on structured light sensing and point triangulation. The more experienced reader may skip ahead to the novel motion compensation strategy, which is described in Section 2.4.

### 2.1. Background

In the following, we give a short introduction to the working principles and characteristics of structured light scanning.

Structured light involves a projector shining one or more codified light patterns on the object surface. The geometry of the object deforms the pattern and the resulting scene is captured by one or more cameras. The projector can be seen as an inverse camera and the patterns help to establish distinct and accurate sub-pixel camera-projector correspondences. This enables structured light sensors to potentially achieve $\mu$m-mm accuracy, even when scanning smooth, textureless surfaces.

There are two main limitations of structured light scanning.

First, typical structured systems do not work well when measuring reflective surfaces. This is because the lack of any light absorption or scattering on the reflective surface may cause regions of the camera image to become saturated. Nevertheless, structured light imaging has proven its effectiveness in scanning a variety of materials that could be found at a construction site, such as brick, sand, rock, cement, wood, marble, bright coloured paint and dark tree bark [11–14].

Second, structured light systems do not work well outdoors where sunlight may overpower the projected light [15]. However, the intended usage environments of the SL sensor are indoors where lighting can be controlled in a sufficient degree.

As such, we believe that structured light scanning is a highly suitable sensing method for the purposes of obtaining high-quality scans of construction surfaces indoors.

### 2.2. Phase shift profilometry

Over the years, numerous light patterns that have been developed [16]. Out of these, the PSP codification technique is distinctive because of its ability to 1) establish sub-pixel accurate camera-projector correspondences 2) produce scans of high spatial resolution, which is necessary to detect minute details on construction surfaces 3) achieve high scanning speeds using hardware triggering 4) tolerate some lens defocusing and hence can work over a larger scanning range [17]. Moreover, experiments from [11] has demonstrated that PSP structured light scanning works well on materials commonly found on construction sites. One of the most straightforward PSP patterns with low computational complexity is the 3 + 3 pattern [17]. As suggested by its name, the 3 + 3 method involves displaying two sets of projections, each containing 3 patterns. The first set of projections include 3 high frequency sinusoidal patterns, each shifted by a phase of $2\pi/3$ rad. The resulting images $I_i$, $i = 1,2,3$ can be modelled as

$$I_i(u,v) = A(u,v) + B(u,v)cos(\phi(u,v) - 2\pi(i-1)/3) \quad (1)$$

where $A(u,v)$ is the average intensity, $B(u,v)$ is the modulation intensity while $\phi(u,v)$ is the wrapped phase. The wrapped phase can be solved using the equation below. $(u,v)$ is omitted from the terms $\phi$, $I_1$, $I_2$ and $I_3$ for brevity.

$$\phi = \operatorname{atan2}\left(\sqrt{3}(I_2 - I_3), 2I_1 - I_2 - I_3\right) mod\ 2\pi \quad (2)$$

where atan2$(y,x)$ is the four-quadrant inverse tangent function. The modulo operation over $2\pi$ wraps phase values ranges over [0 - $2\pi$), as introduced by [18,19].

The phase information from the high frequency pattern $\phi_h$ contains several $2\pi$ wraps and it is difficult to differentiate between the multiple fringes. The process of distinguishing these phase ambiguities is termed as phase unwrapping. While numerous phase unwrapping methods have been proposed [20], the most straightforward way of solving this issue is to use the standard temporal phase unwrapping (TPU) technique. This method recovers the absolute phase by projecting another 3 phase-shifted sinusoidal patterns of unit frequency and using them to compute $\phi_l$ using the resulting images. Note that these 3 unit frequency projections are the second set of patterns in the 3 + 3 pattern. The absolute phase $\Phi(u,v)$ can then be computed using the equations

$$\Phi(u,v) = \phi_h(u,v) + 2\pi k(u,v), k(u,v) \in \mathbb{Z}\ \cap\ [0, N_{fringes} - 1] \quad (3)$$

$$k(u,v) = round\left(\frac{N_{fringe} \cdot \phi_l(u,v) - \phi_h(u,v)}{2\pi}\right) \quad (4)$$

where $N_{fringe}$ is the number of fringes in the high frequency patterns and $round(x)$ rounds the value $x$ to the closest integer. As a final step, we convert the absolute phase to its corresponding horizontal or vertical projector pixel coordinate, based on whether the sinusoidal pattern propagates along the projector's x or y axis respectively

$$p(u,v) = \frac{\lambda}{2\pi}\Phi(u,v) \quad (5)$$

where $\lambda$ is the fringe wavelength in projector pixels. The interested reader can refer to [17] for more information regarding the standard TPU algorithm.

### 2.3. Point triangulation

In the field of PSP, there are two ways of obtaining depth information. The first is the phase-height method [21,22] where a reference plane is defined and researchers try to map the phase difference between the object and the reference plane to the object's height (with respect to the reference plane) using various mathematical models [23–26]. Well-known models used for phase-height mapping include the linear [27], linear inverse [28] and polynomial [29,30] phase-height models. The second method is the triangular stereo method whereby the camera and projector are modelled as a stereo pair. Camera-projector correspondences are then triangulated using projective geometry to obtain a 3D point in space. We chose the latter approach because it directly measures geometry with respect to the robot instead of a reference plane and also because projective geometry is widely used in various fields of robotic perception such as visual odometry [31] and structure from motion [32].

In our case, the pinhole camera model is applied to both the camera and projector. The pinhole camera model maps 3D points in the world coordinate frame $(x^w, y^w, z^w)$ to 2D pixel coordinates $(u,v)$. This gives the

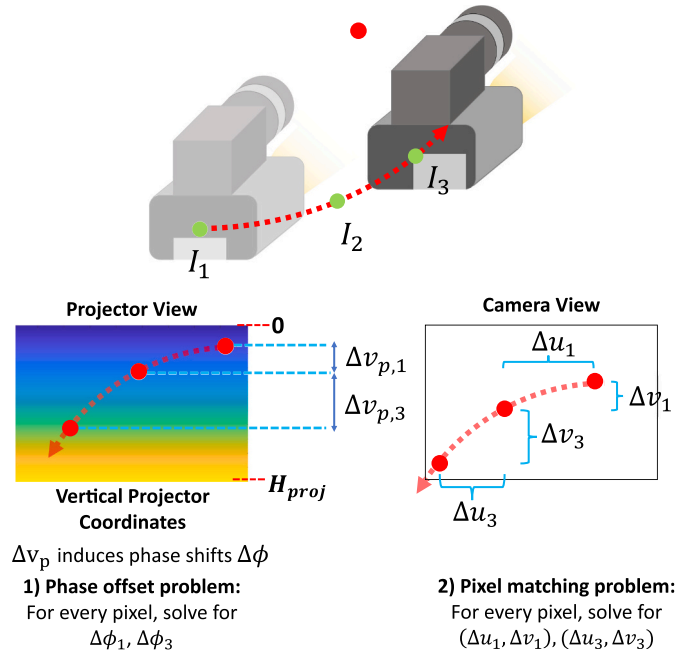

**Fig. 1.** Diagram illustrating the two problems that are required to perform motion compensation for a single 3D point in space (i.e. the red dot). Green dots show the location of captured images. A 3-step PSP vertical pattern is projected and the reference image is $I_2$. (For interpretation of the references to colour in this figure legend, the reader is referred to the web version of this article.)

equations

$$s_c \begin{bmatrix} u_c \\ v_c \\ 1 \end{bmatrix} = K_c [R_c | t_c] \begin{bmatrix} x_w \\ y_w \\ z_w \\ 1 \end{bmatrix} = M_c \begin{bmatrix} x_w \\ y_w \\ z_w \\ 1 \end{bmatrix} \quad (6)$$

$$s_p \begin{bmatrix} u_p \\ v_p \\ 1 \end{bmatrix} = K_p [R_p | t_p] \begin{bmatrix} x_w \\ y_w \\ z_w \\ 1 \end{bmatrix} = M_p \begin{bmatrix} x_w \\ y_w \\ z_w \\ 1 \end{bmatrix} \quad (7)$$

$s$ is the scaling factor and $K$ is the $3 \times 3$ intrinsic matrix. $R$ is the $3 \times 3$ rotational matrix and $t \in \mathbb{R}^3$ is the translation vector that describes the orientation and position of the world frame respectively, with respect to the camera frame. $[R|t]$ is known as the extrinsic matrix. We define the world frame to coincide with the camera frame. Therefore, $R_c$ is an identity matrix and $t_c$ is a zero column vector. Given camera pixel coordinate $(u_c, v_c)$, its corresponding $u_p$ or $v_p$ computed in the previous subsection and the projection matrices ($M_c$ and $M_p$), we can establish 5 equations with 5 unknowns ($s_c$, $s_p$, $x_w$, $y_w$, $z_w$). Solving this system of equations algebraically [21,33] provides us with the 3D triangulated point ($x_w$, $y_w$, $z_w$).

### 2.4. Motion compensation under linear motion

#### 2.4.1. Problem formulation

Many PSP patterns including the 3 + 3 pattern sequence require multiple projections, making them sensitive to motion. Moving either the measured surface or the sensor would result in motion ripples in the final point cloud. There are two main approaches to enable multi-shot PSP methods to work during motion: 1) project the patterns at high speeds such that the effects of motion are negligible [34] 2) take into account motion using mathematical modelling and then take measures to compensate, minimise or mitigate its effects. For our method, we perform the latter.

Fig. 1 illustrates the two key problems that have to be solved to successfully perform motion compensation. The first is the pixel matching problem, where we need to know how a 3D point's location in the image shifts over time (estimating $(\Delta u, \Delta v)$). The second is the phase offset problem, where the movement introduces a phase error $\Delta \phi$ because the projector coordinates in the propagation direction of the sinusoidal pattern of a single 3D point varies over time. For a phase-shifting PSP pattern with N steps, the i-th image ($i = 1, …, N$) in the sequence is modelled as

$$I_i(u + \Delta u_i(u, v), c + \Delta v_i(u, v)) = A + B\cos(\phi(u, v) + \Delta \phi_i(u, v) - 2\pi(i-1)/N) \quad (8)$$

Solving the full motion compensation problem is no trivial task because, as seen from eq. 8, it involves estimating $\Delta u$, $\Delta v$ and $\Delta \phi$ for all camera pixels and across all images.

#### 2.4.2. Proposed novel motion compensation strategy

To simplify the motion compensation problem, we first limit the motion to be linear about a single axis with no change in orientation, in the direction perpendicular to the direction of the sinusoidal pattern. By doing so, the encoded projector pixel coordinate of a 3D point is independent of the linear movement and hence no phase offset is induced [35].

For the pixel matching problem, most existing works model the changes in the images over time to be a 2D rigid transformation, which allows majority of the pixels to be matched by simply translating and/or rotating the images. Various methods to find the required 2D transformations have been explored, including the tracking of external markers [36], tracking of visual keypoints [37], object tracking [38] and phase correlation image registration [39]. In our case, we perform phase correlation image registration to find the single direction translations to align the images. Note that [39] performs one image registration between every frame set and then evenly interpolates the image shifts between the individual images, and also did not propose any restrictions in motion. For our proposed method, we perform image registration between the chosen reference image and every other image in the frame set, enabling our method to perform well even in cases of non-uniform motion.

Phase correlation image registration is a computationally efficient, frequency-domain based technique used to align two images first introduced by [40]. It is robust to occlusions and hence works well for our case where the projected sinusoidal patterns cause regions in the scene to be unevenly illuminated over the captured image sequence.

Assume we have two images $f$ and $m$ that are of size $h \times w$ that have a circular translational shift of $(\Delta x, \Delta y)$ between them

$$m(x, y) = f(mod(x - \Delta x, w), mod(y - \Delta y, h)) \quad (9)$$

Let $F$ and $M$ be the discrete Fourier transform (DFT) of $f$ and $m$ respectively. The normalised cross-power spectrum between $F$ and $M$ is given by

$$C(u, v) = \frac{F \circ M^*}{\|F \circ M^*\|} = e^{2i\pi\left(\frac{u\Delta x}{w} + \frac{v\Delta y}{h}\right)} \quad (10)$$

where * denotes the complex conjugate and ∘ is the Hadamard (elementwise) product. The required translation $(\Delta x, \Delta y)$ can then be found by identifying the peak of $C$'s inverse Fourier transform

$$(-\Delta x, -\Delta y) = \underset{x,y}{\mathrm{argmax}}\; c(x, y) = \underset{x,y}{\mathrm{argmax}}\; \delta(x + \Delta x, y + \Delta y) \quad (11)$$

where $\delta$ is the Kronecker delta function.

After solving the pixel matching problem using phase correlation image registration and since no phase error is induced due to the constrained motion, the aligned images can then be processed using standard PSP algorithms to obtain a point cloud in which the effects of motion are minimised.

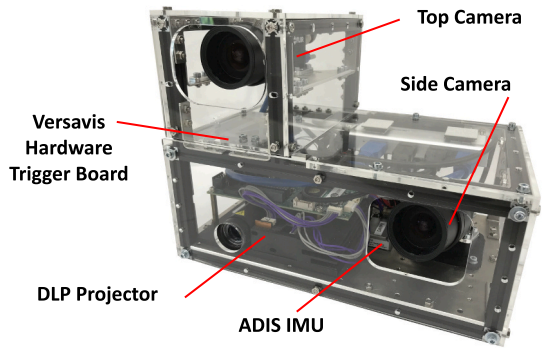

**Fig. 2.** Labelled diagram of SL Sensor.

triggered at 30 Hz while the projector is triggered at 5 Hz and set to project an entire pattern sequence after each trigger. Each pattern is exposed twice to ensure that each image will capture the full projection of a single pattern[2] despite any inherent image capture latencies.

On the computer's end, projector trigger timings published over ROS by the Versavis node to identify the received images. Given that we receive a projector trigger timing for the $i$th pattern at time $t^p$ and an image timestamped $t^c$, we identify it as the matching image $I_i$ if it falls within the time range

$$-\Delta t_{tol,l} \leq t_c - \left(t_p + (i-1)\Delta t_{img}\right) \leq \Delta t_{tol,u} \qquad (12)$$

where $\Delta t_{img}$ is the time interval between camera triggers and setting $\Delta t_{tol,l} = \Delta t_{tol,u} = 0.2 \times \Delta t_{img}$ works well for our system.

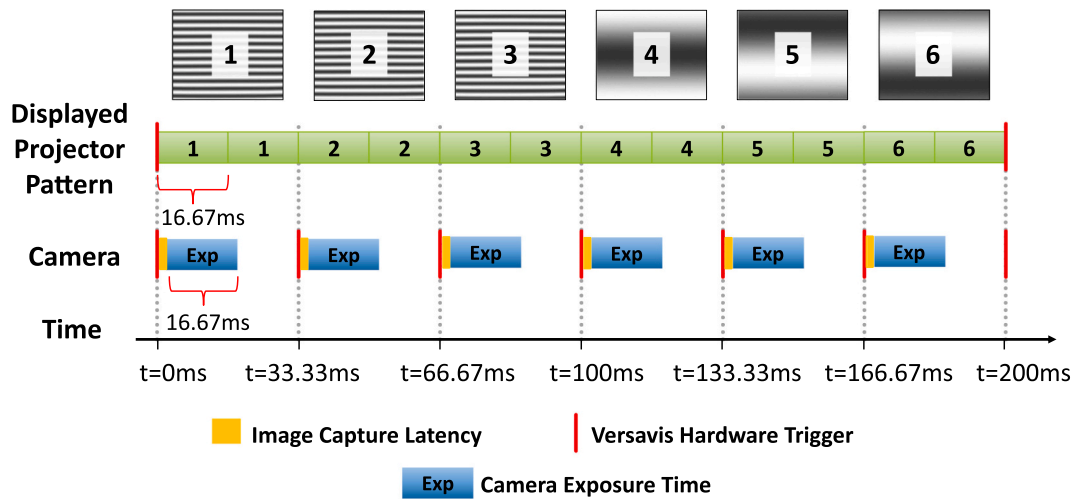

**Fig. 3.** Triggering schedule that allows the SL sensor to produce scans at 5 Hz.

## 3. Implementation

### 3.1. Hardware

The built SL Sensor is shown in Fig. 2. It contains two industrial complementary metal oxide semiconductor (CMOS) colour cameras (1440 × 1080 resolution) and a Digital Light Processing (DLP) projector (912 × 1140 resolution). The camera and projector are hardware triggered by a Versavis board [41] that facilitates the synchronisation of pattern projection and image acquisition. The SL Sensor also has an IMU that was not used for the proposed algorithms but can be used in the future for SL motion compensation strategies where a motion estimate is required [42], e.g. visual-inertial odometry. The cameras and projector are positioned such that their field of views overlap when scanning an object placed 0.3–1.0 m away from the sensor. In this configuration, the SL Sensor has a size of 24.6 × 15.8 × 18.8 cm. Note that the sensor can be made more compact by decreasing the camera-projector baseline, at the expense of greater depth uncertainty [43]. For the camera lens, we chose to use a varifocal lens so that the focus can be fine-tuned to our specific scanning range. Furthermore, the chosen camera lens has an adjustable iris. This enables the user to adjust the aperture based on the surface's colour and material, to ensure that the captured images are not overexposed.

### 3.2. Hardware triggering

In order to achieve high speed scanning, the projector and cameras are hardware triggered by the open-source Versavis board and the triggering schedule for a single scan is shown in Fig. 3. The cameras are

### 3.3. Ensuring a linear intensity response

The PSP decoding process assumes that the intensity of the projected light and the intensity measured by the camera have a linear relationship. To achieve this, all camera image processing options (e.g. gamma correction, automatic gain control, etc) need to be turned off. Furthermore, the projector is set to *pattern mode* where it displays patterns stored in its flash memory without any additional image processing. This is a better alternative to displaying images over HDMI feed where the projector automatically applies gamma correction to the images [45] that needs to be compensated for using additional steps [46,47].

### 3.4. Sensor calibration

Sensor calibration is required to obtain the intrinsics (intrinsic matrix and lens distortion coefficients) as well as extrinsics (transformation between camera and projector). We adopt the well known lens distortion model proposed by Bouguet [48] which consists of 5 coefficients - 3 for radial distortion and 2 for tangential distortion.

Calibration of the SL Sensor is done between each camera-projector pair separately. This is sufficient for our current use case where the depth estimation process only utilises one camera-projector pair at any given time, but it can be extended to a joint callibration sequence if future applications require it. This pairwise calibration is done with the procedure from [49] whereby PSP patterns are projected onto a grey checkerboard calibration target. The computed shading image is used to extract the checkerboard patterns and a local homography method [50] is used to extract the corresponding projector coordinates from the decoded projector coordinate maps. With the checkerboard coordinates from both the camera and projector, the intrinsics and extrinsics of the two devices are estimated using OpenCV's Camera Calibration and 3D Reconstruction library [51].

---

[2] Note that a full projection needs to be captured because DLP projectors display grayscale patterns using binary pulsewidth modulation [44].

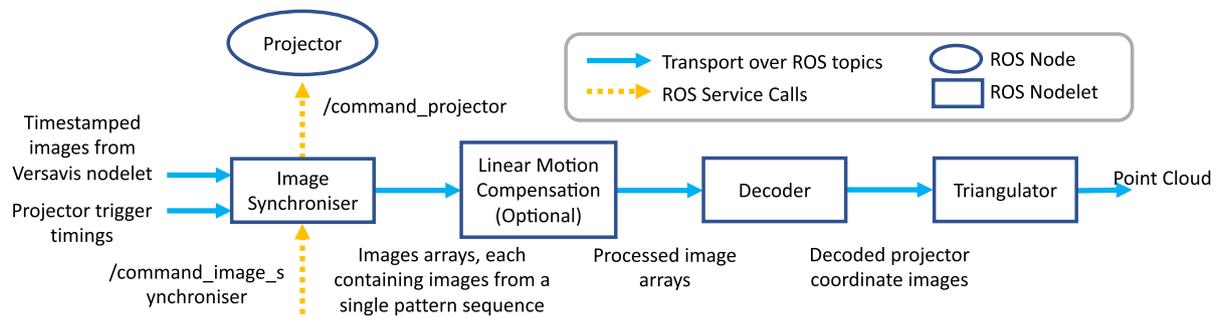

**Fig. 4.** Block Diagram of SL Sensor's Software System.

The complete calibration steps for SL Sensor is as follows:

1. Take images of the calibration board using the primary camera
2. Perform pairwise calibration to get the primary camera intrinsics, projector intrinsics and primary camera-projector extrinsics
3. Take images of the calibration board using the secondary camera
4. Perform pairwise calibration to get the secondary camera intrinsics and secondary camera-projector extrinsics while fixing the projector intrinsics to those estimated in step 2

In our case, the primary camera refers to the top camera while the secondary camera is the left camera.

*3.5. Software*

The SL Sensor software is written under the Robot Operating System (ROS) framework. This ensures that it can be easily used with or integrated into existing robotic packages and solutions contributed by the ROS user community.

The main reconstruction pipeline is broken down into 4 main ROS nodelets (Fig. 4). The usage of nodelets enables efficient zero copy pointer passing of images between subprocesses while still ensuring the modularity of pipeline. Parts of the Decoder and Triangulator nodelets adapt code from SLStudio software package [9].

1. The **Image Synchroniser** nodelet takes in timestamped images as well as projector trigger timings from the Versavis ROS nodelets and based on eq. 12, groups images that belong to the same pattern sequence into a single image array for downstream processing. To enable the image synchroniser, the user must first send a service call to indicate the pattern sequence to be projected and the number of scans required. It will in turn send a service call to the Projector node to initialise the flashing of patterns by the Lightcrafter 4500 before the image grouping process is started.
2. The **Linear Motion Compensation** nodelet performs phase correlation image alignment on the image array as detailed in Section 2.4.2. If not required, the reconstruction pipeline can be initialised without it.
3. The **Decoder** nodelet receives the captured images and converts them into horizontal and/or vertical projector coordinate maps as specified in Section 2.2.
4. The **Triangulator** nodelet receives the decoded projector coordinate maps and uses them to generate the final point cloud. To ensure real-time performance, the computations stated in Section 2.3 are sped up using a pre-computed determinant tensor [52].

**4. Experimental evaluation**

In the following, we first validate that the open source sensor system described in Section 3 can deliver the quality expected from a multi-projection structured light system, and that such increased precision compared to standard robotic sensors enables novel sensing applications for construction robotics.

In a next step, we evaluate our novel motion compensation strategy described in Section 2.4.2 in a controlled lab experiment.

Finally, we conduct a larger real-world experiment on a construction site, validating that the increased precision of multi-projection structured light in combination with our motion compensation strategy can be used to scan large surfaces with a high level of detail.

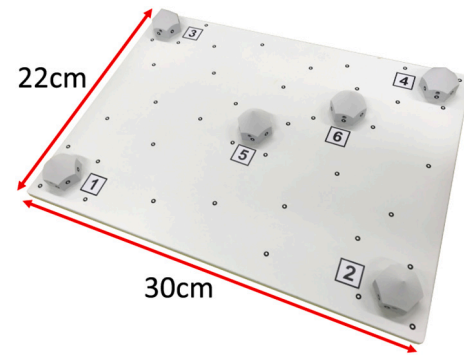

**Fig. 5.** The custom evaluation board used for the cone fitting accuracy test.

*4.1. Static tests*

We compared the SL Sensor with with two commercial-grade depth cameras - the Azure Kinect and the RealSense L515. These depth cameras are commonly used in mobile robotic applications for depth sensing. Note that for all experiments in this subsection, all the sensors were positioned around 50 cm away from the scanned objects and the scans were performed indoors where the temperature is regulated to be around 20°C. This is to ensure that the tests were done within the reported scanning range and operational temperature range of the evaluated sensors.

To evaluate sensor accuracy, a custom evaluation board (Fig. 5) was scanned. The board consists of 6 metallic cones mounted on a glass sheet, coated with a fine layer of non-reflective white paint. Cone fitting was performed from the 3D point clouds (10 scans per sensor) and the distances between Cone 1 and all other cones were computed. The ground truth distances were obtained using a metrology-grade GOM ATOS Core 300 structured light scanner that has an accuracy of approximately 10 μm. The results of the cone fitting test are shown in Table 1. For the SL Sensor, the distances computed deviated from the ground truth by less than 1 mm, with the exception of the 1 → 4 distance measured from the left camera. In general, the left camera performed slightly worse than the top camera. This is mainly because the orientation of the camera resulted in some of the cones to be only partially scanned (Fig. 6). Nevertheless, the SL Sensor outperformed the other two sensors in terms of both accuracy and measurement uncertainty.

**Table 1**
Statistics of the measured distance between the fitted cones from the scans for the various sensors across 10 scans.

| Distance | SL Sensor (Top Cam) | | SL Sensor (Left Cam) | | RealSense L515 | | Azure Kinect | |
|---|---|---|---|---|---|---|---|---|
| | RMSE/mm | Std Dev/mm | RMSE/mm | Std Dev/mm | RMSE/mm | Std Dev/mm | RMSE/mm | Std Dev/mm |
| 1 → 2 | **0.125** | **0.041** | 0.701 | 0.111 | 5.546 | 2.694 | 2.168 | 0.258 |
| 1 → 3 | 0.381 | **0.062** | **0.328** | 0.091 | 2.763 | 1.487 | 1.452 | 0.324 |
| 1 → 4 | **0.713** | **0.042** | 1.144 | 0.399 | 2.875 | 1.625 | 2.674 | 0.246 |
| 1 → 5 | **0.077** | **0.045** | 0.336 | 0.102 | 4.292 | 2.129 | 2.041 | 0.253 |
| 1 → 6 | **0.246** | **0.054** | 0.501 | 0.187 | 2.165 | 1.311 | 3.197 | 0.244 |

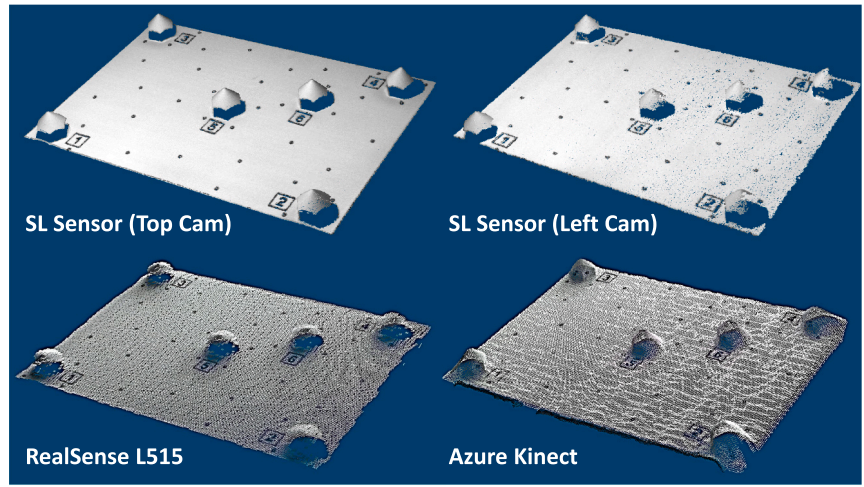

**Fig. 6.** Point clouds obtained from the evaluated sensors of the custom evaluation board.

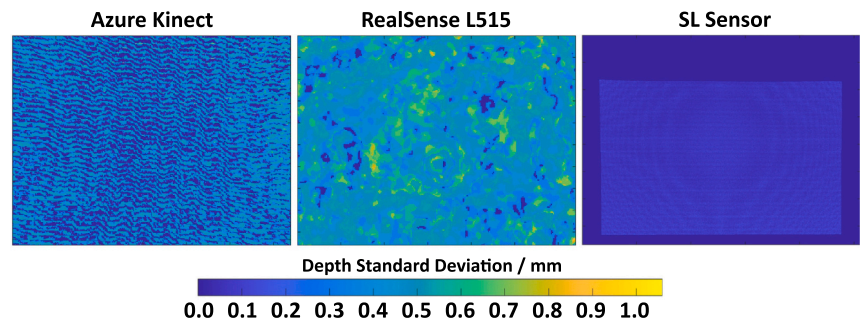

**Fig. 7.** Plots of the depth standard deviation of each pixel in each device.

To evaluate sensor precision, we took 10 consecutive static scans of a flat, white, matt surface for each sensor and analysed how the measured depths varied for each pixel over the scans. This is similar to the tests done in [53–55]. The results from the precision tests are plotted in Fig. 7. The Azure Kinect's plot has a wavy pattern where half of the pixels have close to zero measurement uncertainly while the other half experienced some variations in readings (about 0.3 mm in depth standard deviation) over consecutive scans. The RealSense L515 performed worse, with several spots in the image registering depth standard deviation values of over 0.7 mm. The SL sensor performed the best, as clearly seen from its predominantly dark blue plot. Notice that there is a dark border surrounding the SL sensor's plot. These are the regions of the image where no projected light was observed and hence they were not evaluated for this test. For all three plots, we computed the mean and maximum value of the depth standard deviation over all pixels that had valid depth information (Table. 2). The SL sensor reported the smallest mean and maximum values that are almost four times lower than the values obtained from the Azure Kinect and Realsense L515. Hence, we can conclude that the SL sensor is the most precise among the sensors tested.

**Table 2**
Table showing the mean and maximum values of the depth standard deviation over all valid pixels of the three devices.

| Device name | Depth Std. Dev./mm | |
|---|---|---|
| | Mean | Max |
| Azure Kinect | 0.256 | 1.059 |
| RealSense L515 | 0.470 | 0.996 |
| SL Sensor | **0.070** | **0.251** |

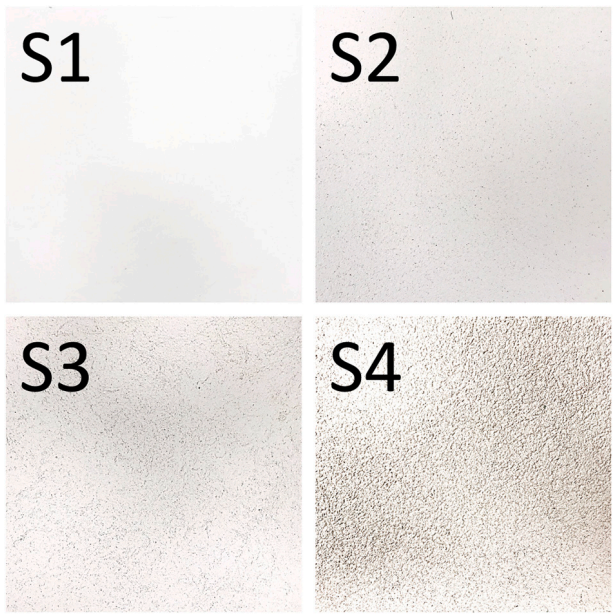

**Fig. 8.** Pictures of the surface quality samples that were measured.

**Table 3**
Table showing the ESD values obtained for the four gypsum plaster samples. The ESD values for the GOM ATOS Core 300 and Helios Lucid are those reported in [58].

| Sensor | ESD/mm | | | |
|---|---|---|---|---|
| | S1 | S2 | S3 | S4 |
| GOM ATOS Core 300 | 0.008 | 0.026 | 0.053 | 0.146 |
| SL Sensor | 0.076 | 0.080 | 0.099 | 0.155 |
| Lucid Helios | 0.91 | 0.88 | 0.87 | 0.94 |

In summary, while the Azure Kinect and the Realsense L515 have significantly larger scanning ranges (0.5–5.46 m for the Azure Kinect [56] and 0.25–9.0 m for the Realsense L515 [57]), the experiments confirm that multi-projection structured light scanning for surface interaction as used by SL Sensor is superior to these commercial-grade sensors in terms of precision and accuracy in the intended range band. We conclude that our open-source SL Sensor works as expected.

*4.2. Scanning of surface quality samples*

Surface roughness is a key metric in determining the surface finish quality of fabricated structures in a construction site. To determine SL sensor's potential to classify and assess surface roughness, we used it to scan four 17 × 17 cm specimens of gypsum plaster (Fig. 8), which is a commonly used surface finishing material for building interiors. The plaster samples were placed 50 cm away from the sensor during scanning. 30cm$^2$ patches of the scans were then used to compute the empirical standard deviations (ESD) of orthogonal distances from the best fit planes of each extracted patch. We compare these values with the ESD values reported in [58] that measured the same samples, but with a GOM ATOS Core 300 structured light scanner (ground truth) and a Lucid Helios time-of-flight (ToF) depth camera. As seen from the results in Table 3, while there is no discernible difference in ESD values across all samples for the Helios, the SL Sensor was able to pick up on the trend of increasing roughness across the samples as confirmed by the ATOS Core scanner.

*4.3. Scanning along a linear rail*

Our motion compensation strategy was first tested on a single pattern sequence where we took images of a white mask at a distance of 50 cm. The top camera was used for the scanning and the vertical 3 + 3 pattern was projected. The sensor was shifted linearly to the right by 2 mm after every image. Fig. 9 shows the scans produced with and without motion compensation. It is clear that our motion compensation strategy is able to remove the distortions caused by the linear motion. We also took a cross section of the resulting depth images at the location indicated by Fig. 10. The cross section plot further affirms the effectiveness of our motion compensation strategy as it is able to remove the 1-2 mm deep motion ripples found on the mask's forehead.

*4.4. Scanning of a spray plastered wall with adaptive projections*

Motion compensation enables scanning of structures that are larger than the sensor's field of view, by moving the scanner along the structure and registering individual scans together.

To demonstrate this capability of SL Sensor in an actual construction robotics task, we used it to scan segments of a spray plastered wall fabricated using the methodology described in [59]. We attached the SL Sensor to a MABI Speedy 12 robotic arm which moved the sensor along a straight line trajectory at a speed of around 0.0125 m/s while maintaining a sensor-wall distance of 0.3–0.5 m during the scans. The sensor scanned two regions of the indoor fabricated workpiece shown in Fig. 11. We utilised the adaptive nature of our sensor and combined a horizontal and vertical trajectory over each region, switching the projected pattern and the camera used for triangulation in between these two motions. For each region, we performed a 50 cm horizontal and 30 cm vertical pass with motion compensated 3 + 3 pattern 3D scanning. The individual scans were then merged together using pairwise point-to-plane iterative closest point algorithm (ICP) from the *libpointmatcher* library [60]. For comparison, we obtained scans of the two regions using a Leica Nova MS50 laser scanner (referred to as TLS in this paper) that has an accuracy of 2 mm. Point clouds from both the TLS and SL Sensor were then converted into meshes using Poisson surface reconstruction [61]. While the precision of the TLS is lower than individual SL Sensor scans, it allows us to quantify deviations of accumulated error due to point cloud registration.

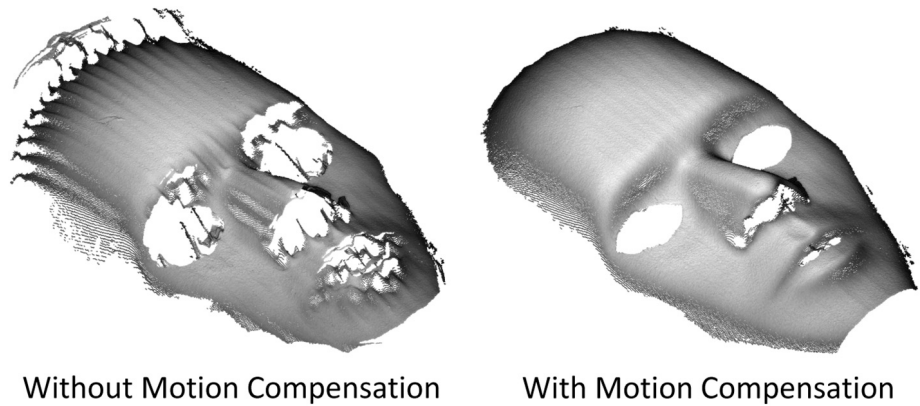

**Fig. 9.** Images of the 3D point clouds with (right) and without (left) motion compensation of the scanned mask.

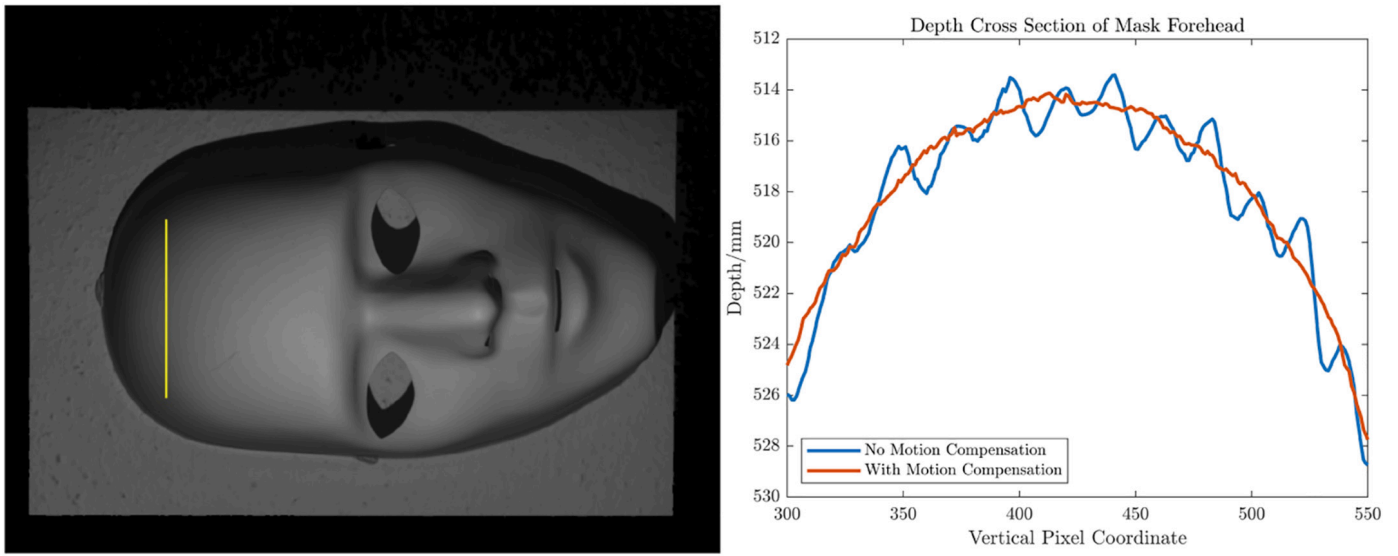

**Fig. 10.** Left: Image of the scanned mask and the yellow line shows the location of the cross section Right: Cross section plot at the forehead region of the mask scans. (For interpretation of the references to colour in this figure legend, the reader is referred to the web version of this article.)

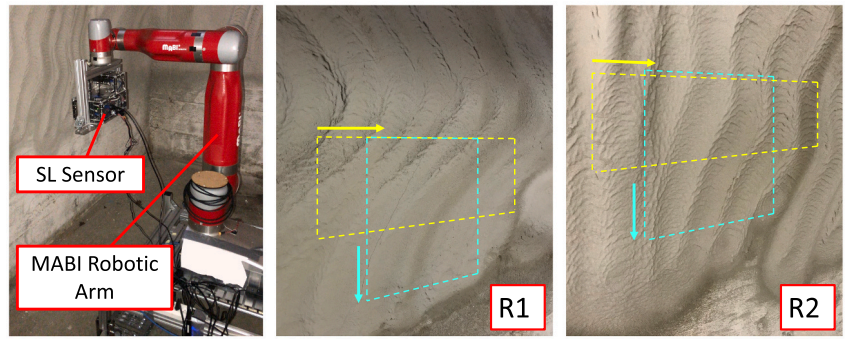

**Fig. 11.** Left: Image of the experimental setup. Right: Images of the two regions of the scanned spray plastered wall. The yellow and cyan boxes demarcate the areas scanned for the 50 cm horizontal pass and 30 cm vertical pass respectively. The arrows indicate the direction of scanning. (For interpretation of the references to colour in this figure legend, the reader is referred to the web version of this article.)

To compare the meshes from the TLS and SL Sensor, we first aligned them together using point-to-point ICP. Next, we computed the cloud-to-mesh (C2M) distance between the vertex points of the SL Sensor mesh and the TLS mesh using the 3D point cloud processing software *CloudCompare* [62].

The resulting meshes and C2M distance histograms are shown in Fig. 12 and Fig. 13. Note that for the heat map of C2M distances, points that deviate by less than 1 mm from the TLS mesh are coloured white. Visually, the meshes from the SL Sensor match well with those produced by the TLS. More quantitative results are presented in Table 4. The table reveals that 99% of SL Sensor mesh vertices deviate by less than 2 mm and 3 mm from the TLS mesh for the vertical and horizontal passes respectively. A slightly larger amount of deviation is to be expected for the horizontal passes since there would be greater error accumulation from the larger number of pairwise point cloud registrations performed. This is visually shown by the deviation heat map of the R1 horizontal pass where the left portion of SL Sensor mesh matches well with the TLS mesh but we observe larger deviations on the right where the point clouds were registered last.

Another reason for large deviations between the SL Sensor and TLS meshes is the fact that the SL Sensor was able to capture minor details that the TLS could not. An example of this is shown in the A and B regions (Fig. 13) of the R2 horizontal pass. As shown in Fig. 14, the SL Sensor mesh successfully reproduced the minor edges and ridges that appear in the photos of the actual workpiece as compared to the TLS mesh where the entire area is smooth. The suspected cause of this inaccuracy in the TLS mesh is the fact that these surfaces were orientated away from the laser emitter of the TLS due to the undulating nature of the workpiece, and hence only a sparse point cloud of these regions could be obtained, resulting in the reduction of details captured.

## 5. Conclusion

In this work we introduced a new open-source structured light scanning solution. In contrast to existing sensor solutions, our SL Sensor integrates with existing robotic software over the ROS middleware framework to enable its adaptation to customised 3D scanning procedures. We described our software architecture, hardware setup and calibration procedure and verified that the sensor achieves sub-mm accuracy. We compared it to commonly used sensors in robotic applications as well as commercial high-precision scanners, concluding that our sensor reaches sufficient accuracy for detailed construction applications. We further validated the effectiveness of our novel motion-compensation strategy enabling high-precision PSP scanning under linear motion, and showcased our sensor's ability to switch between multiple patterns dependent on the intended robot motion in a real-world construction setting.

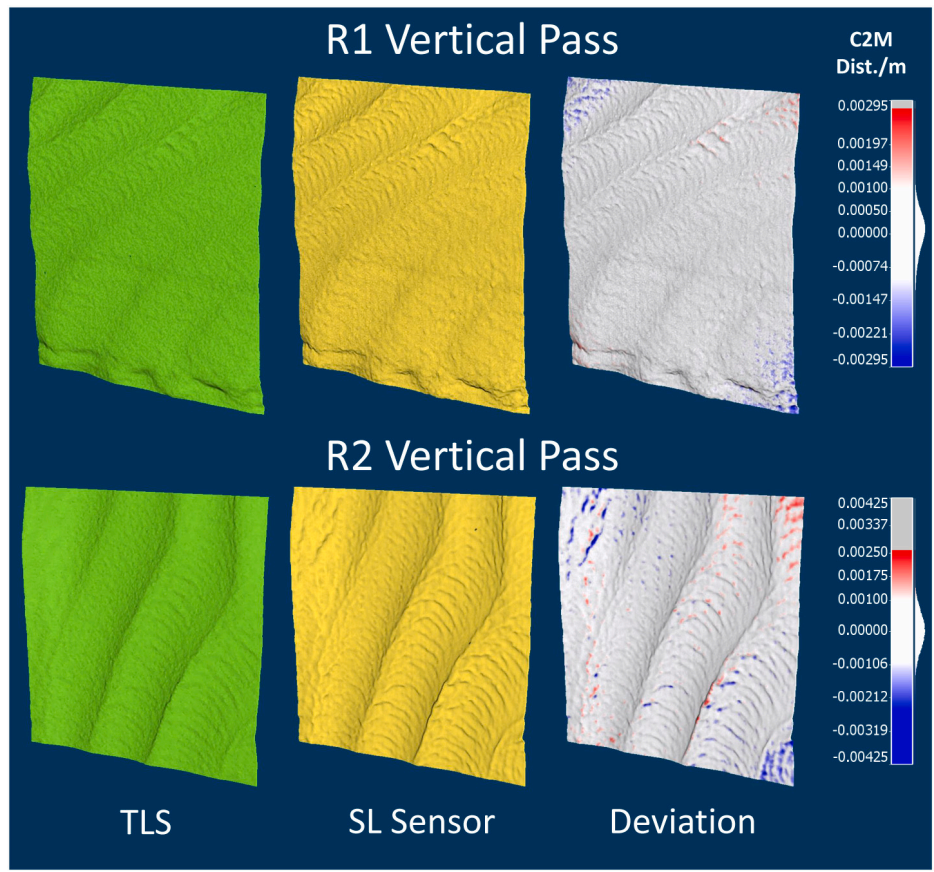

**Fig. 12.** Comparison of the meshes generated by the TLS and SL Sensor from the vertical pass experiments.

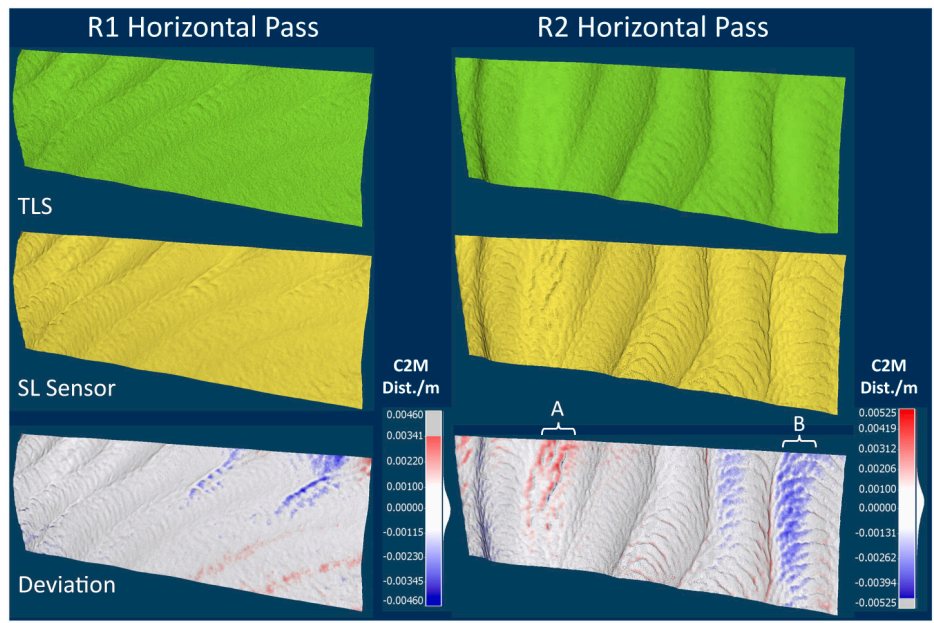

**Fig. 13.** Comparison of the meshes generated by the TLS and SL Sensor from the horizontal pass experiments.

Future work would extend the SL Sensor's capabilities of scanning during linear motion to arbitrary 6 DoF movements. Possible solutions could include a more robust motion compensation strategy or an adaptive pattern projection approach where we use PSP patterns when the sensor is static and switch to another pattern that is more motion tolerant when motion is detected. In addition, multi-way point cloud registration strategies can be explored to reduce error accumulation over scans and ultimately lead to a more accurate merged point cloud.

Table 4
Statistics of the C2M distances computed between the TLS and SL Sensor meshes for the four scans of the spray plastered wall.

| Scan name | Mean/mm | Std. Dev/mm | 0.5 Percentile C2M Distance/mm | 99.5 Percentile C2M Distance/mm |
| --- | --- | --- | --- | --- |
| R1 Vertical | 0.01 | 0.53 | −1.63 | 1.36 |
| R2 Vertical | 0.01 | 0.65 | −1.94 | 1.65 |
| R1 Horizontal | 0.07 | 0.71 | −2.43 | 1.83 |
| R1 Horizontal | 0.1 | 1.06 | −2.99 | 2.87 |

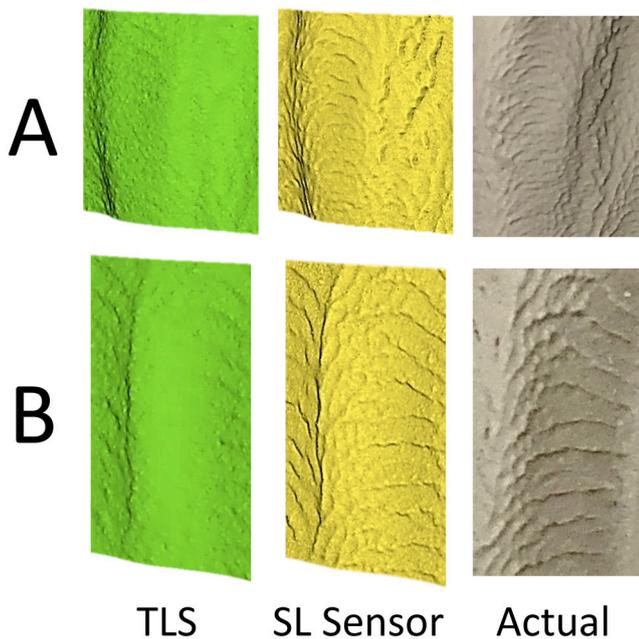

**Fig. 14.** Close up view of the A and B regions labelled in Fig. 13.

**Funding**

This research was proposed and partially funded by an ETH Career Seed grant [grant no SEED-14 20-2]. It furthermore received funding from the HILTI group for research in accurate mobile construction robotics for cement polishing, and the European Union H2020 program under project PILOTING [grant no H2020-ICT-2019-2 871542].

**Declaration of Competing Interest**

The authors declare that they have no known competing financial interests or personal relationships that could have appeared to influence the work reported in this paper.

**Acknowledgements**

The authors would like to thank Johannes Pankert, Robert Presl, Selen Ercan and Valens Frangez for the support they provided to make the experiments in this paper possible. Appreciation should also be given to Michael Riner-Kuhn who provided technical advice during the building of the SL Sensor.